# Nth Absolute Root Mean Error

**Siddhartha Dhar Choudhury, Shashank Pandey**
SRM Institue of Science and Technology
{siddharthadhar_soumen, shashankpandey_mahendra}@srmuniv.edu.in

## Abstract

Neural network training process takes long time when the size of training data is huge, without the large set of training values the neural network is unable to learn features. This dilemma between time and size of data is often solved using fast GPUs, but we present a better solution for a subset of those problems. To reduce the time for training a regression model using neural network we introduce a loss function called Nth Absolute Root Mean Error (NARME). It helps to train regression models much faster compared to other existing loss functions. Experiments show that in most use cases NARME reduces the required number of epochs to almost one-tenth of that required by other commonly used loss functions, and also achieves great accuracy in the small amount of time in which it was trained.

## 1 Introduction

Regression is a common task in machine learning and involves the estimation of values of a target variable given some set of input features. Unlike classification where the output is the probability of the data belonging to other classes, regression requires result to be precise, sometimes to higher number of decimal places. Neural networks can learn to predict these type of values if provided enough training data and computational resources. Often people do not have access to such computational power, Nth Absolute Root Mean Error (NARME) will help train regression models faster.

NARME is a loss function [5] which uses Nth root of the absolute difference between the prediction from the neural network and the actual values provided in the training data. While other common regression losses work good in solving regression problems, they require huge computational resources to be trained in small amount of time.

Taking nth root of the absolute difference of calculated and real value will make sure the loss remains small from the beginning of training process and will eventually result in a smooth curve (epochs v/s loss). Though the difference between the gradients will be small which means that the gradients will slowly move towards achieving the global minimum during back-propagation but still the training works faster than other regression losses, this is because from the beginning of the training process the loss is small.

The NARME loss function introduces a new parameter $n_t$, and $m_t$ which are the order of roots to use in the function. This varies from problem to problem as will be evident from the experiments. The even orders work better than the odd orders of $n_t$ for most of the problems and for $m_t$ odd orders tend to give better results. Choosing the correct value of the hyper parameter $n_t$ and $m_t$ are based on trial and error basis and there is no fixed value of this hyper parameter which works good for every problem. The choice of the value of this hyper parameter plays a crucial role in the overall training process and the resulting predictions.

When compared with standard loss functions that are used most commonly today for solving regression problems like mean squared error, huber loss, etc NARME performs exceptionally well in small training time using the same set of hyper parameters as those used with the former but less number of epochs. In general the number of epochs required by this loss function is one-tenth of that required by the most common loss functions and in the small time for which it is trained it achieves much better result than other loss functions with accuracy of upto four decimal places in some cases.

Most commonly used error functions produce large losses when the training process starts resulting in large gradients which leads to faster gradient descent in the initial epochs, with time the gradients become smaller, but with NARME the loss starts small right from the first epoch of the training process. Imagine the loss to be a ball going down a hill (gradient descent curve), with the proposed loss function the ball never starts at the top of the hill rather begins its descent from a lower range thus takes less time to roll to the bottom of the hill despite of the small updates in the weights during training.

The key to the NARME function is the nth root of the absolute difference between the values which makes the most of the difference and provides a way to speed up training process of regression models using deep neural networks.

## 2 Commonly used loss functions for regression

### 2.1 Mean Squared Error (MSE)

Mean Squared Error (MSE) [2] calculates average squared difference between the estimated prediction (ŷ) and original output value (y). The output of the function is strictly positive due to the usage of squared difference in the equation. The MSE equation is -

$$L(\hat{y}, y) = \frac{1}{n} \sum_{1}^{n} \sqrt{\hat{y} - y}$$

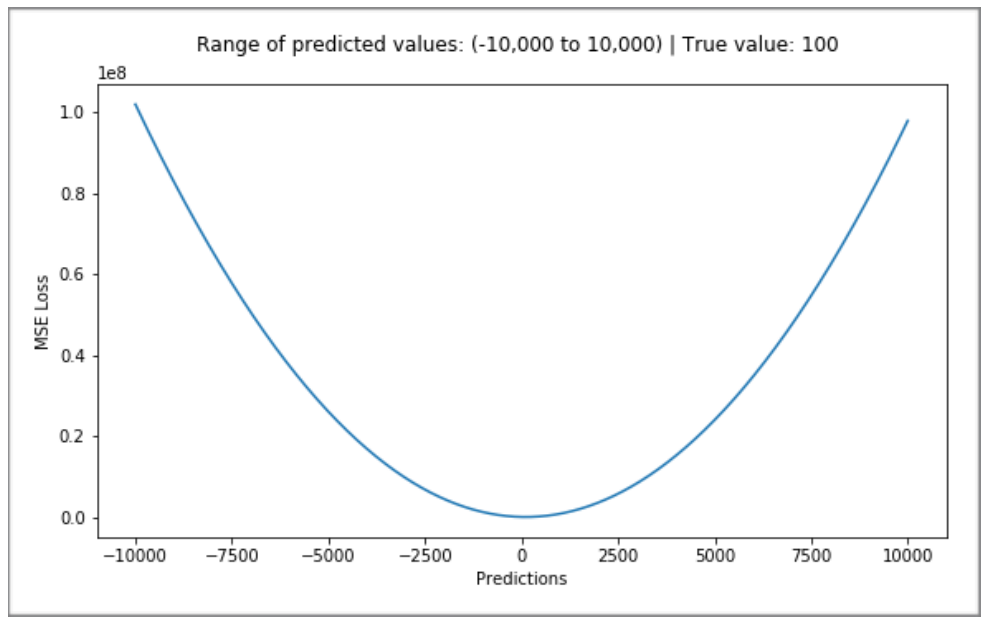

### 2.1 Mean Absolute Error (MAE)

Mean Absolute Error (MAE) calculates sum of absolute differences between the predicted values (ŷ) and output value (y). This measures the average magnitude of errors in a set of predictions, without considering their directions. The equation for MAE is -

$$L(\hat{y}, y) = \frac{1}{n} \sum_{1}^{n} |\hat{y} - y|$$

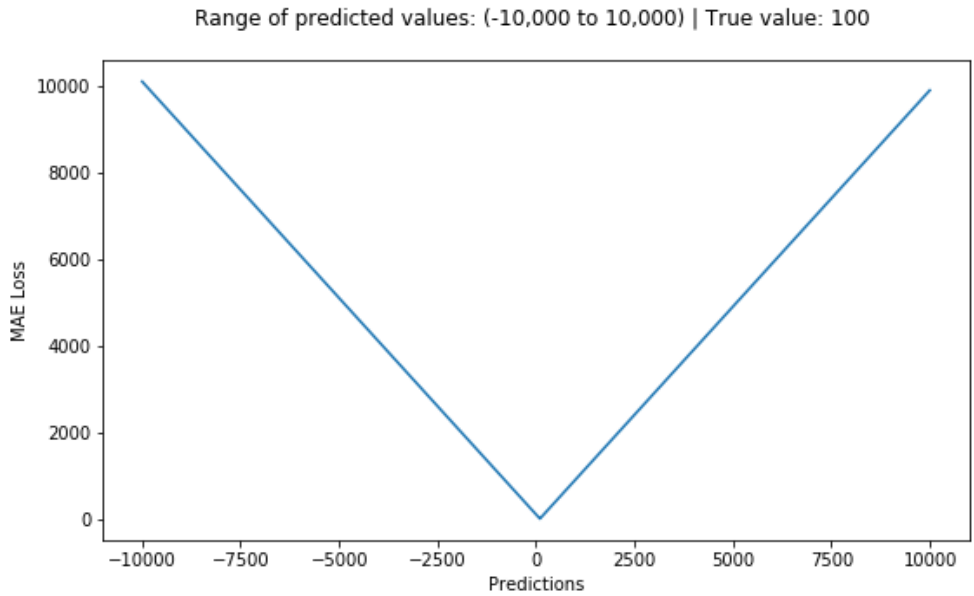

## 2.3 Huber Loss

Huber loss [3] is less sensitive to outliers in data than the squared error loss. It is absolute error, which becomes quadratic when error is small and linear otherwise. It takes a hyper parameter $\delta$ which acts as a threshold value. This hyper parameter plays an important role in the huber loss equation. The equation is -

$$L(\hat{y} - y) = \begin{cases} \frac{1}{2}(\hat{y} - y)^2; & |\hat{y} - y| \leq \delta \\ \delta |\hat{y} - y| - \frac{1}{2}\delta^2; & otherwise \end{cases}$$

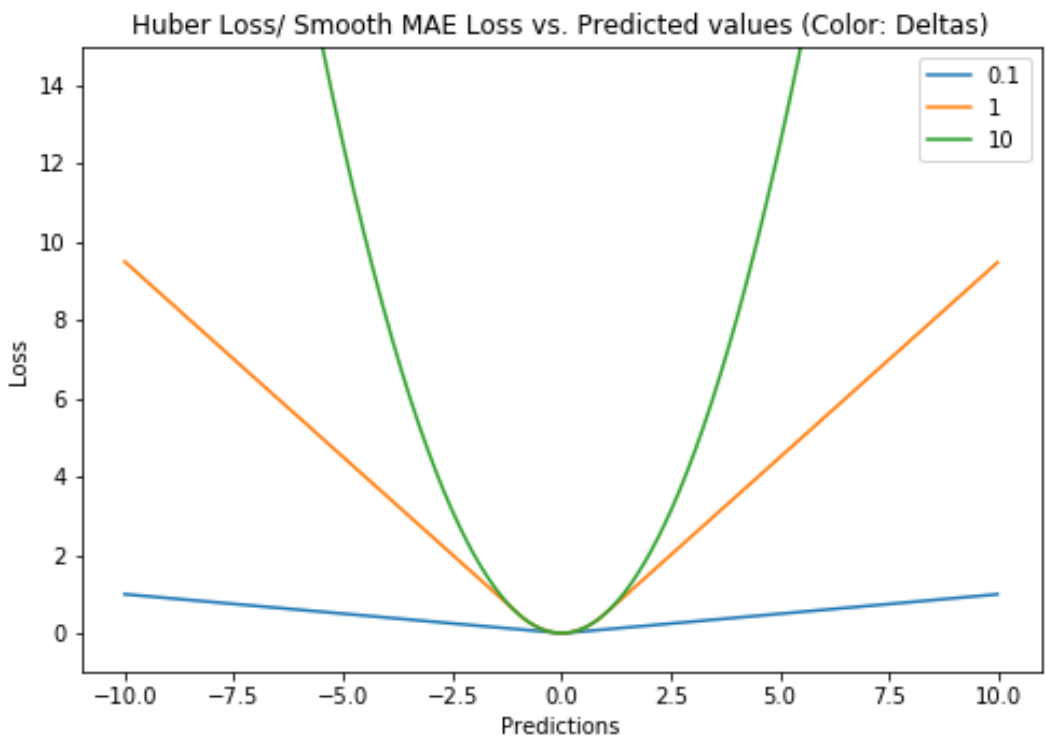

## 2.4 Log-cosh function

This loss function [4] calculates the difference between predicted values ($\hat{y}$) and real output values ($y$), this difference is passed through a hyperbolic cosine function whose result is passed into a log function, this is done for all the training examples and the results are summed up. The log-cosh equation is -

$$L(\hat{y}, y) = \sum_{i}^{n} log(cosh(\hat{y} - y))$$

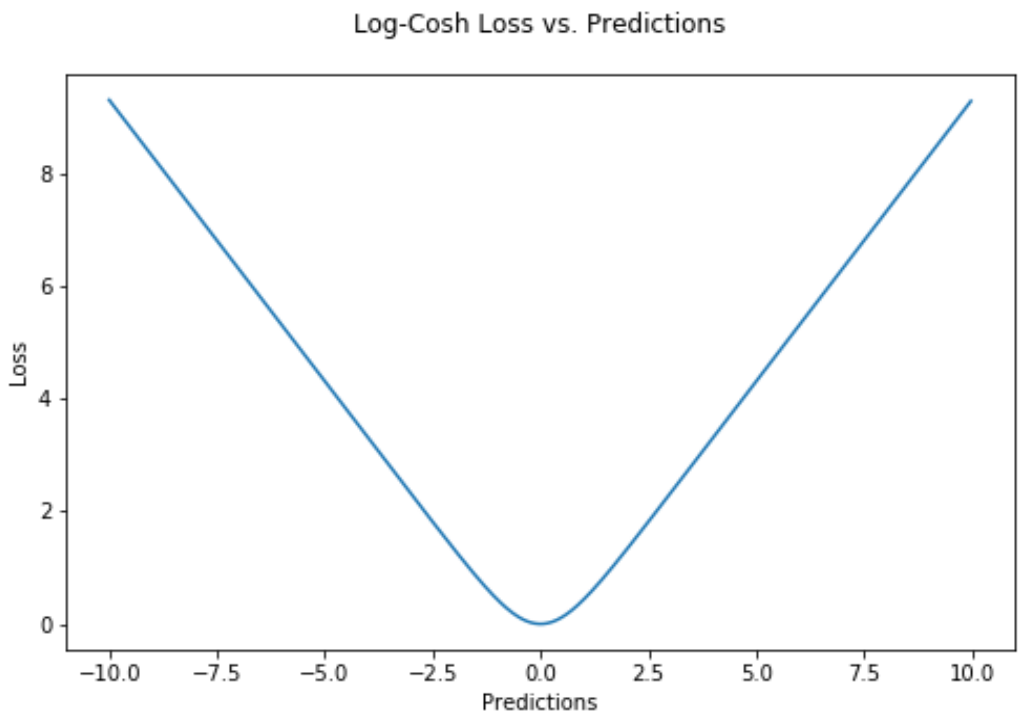

# 3 The NARME Equations

We present two variants of NARME - Single Root (SR-NARME) or Double Root (DR-NARME).

## 3.1 Single Root NARME equation

$$L(\hat{y}, y) = \frac{1}{n} \sum_{1}^{n} \sqrt[n_t]{|\hat{y} - y|}$$

## 3.2 Double Root NARME equation

$$L(\hat{y}, y) = \frac{1}{n} \sum_{1}^{n} max(\sqrt[n_t]{|\hat{y} - y|}, 1) \cdot min(\sqrt[m_t]{|\hat{y} - y|}, 1)$$

The above equation represents the Nth Absolute Root Mean Error functions. It calculates the absolute difference of the predicted value and the actual value and then nth and mth root of the result is calculated, this is done for all the output nodes and mean is taken over all the nodes, thus the name NARME.

The various symbols used in the equation are:-

$\hat{y}$ = Predicted value

$y$ = Actual value

$n$ = Number of output nodes

$n_t$ = Order 'n' of the root

$m_t$ = Order 'm' of the root

The part of the equation that plays major role in it being faster than other values is -

$$\sqrt[n_t]{|\hat{y} - y|} \: and \: \sqrt[m_t]{|\hat{y} - y|}$$

These reduce the loss from the beginning of the training process, the higher the order of the root, the lower the value of the absolute difference of $\hat{y}$ and $y$ will be. Thus $n_t$, $m_t$ and $|\hat{y} - y|$ are inversely proportional to one another -

$$n_t, m_t \propto \frac{1}{|\hat{y} - y|}$$

## 3.3 Relation between order of root and number of epochs

$$n_t \propto \frac{1}{e}$$

The above relation shows that the number of epochs $e$ is inversely proportional to the order of the roots $n_t$ and $m_t$, as the higher the order smaller will be the value of the absolute difference and hence will result in smaller loss from the beginning of training process and can converge faster as compared to lower orders of $n_t$ in the root.

### 3.4 When to use Single Root and Double Root ?

Single root NARME works well for the cases when the training data contains values only in range of one to positive infinity (excluded).

i.e if $x \epsilon [1,\infty)$ then SR-NARME can be used.

When values in the training set has some or all values in the range of 0 and 1 (excluded) then Double Root NARME outperforms Single Root NARME.

i.e if $x \epsilon [0,1)$ then DR-NARME can be used.

### 3.5 Understanding the Double Root NARME

The Double Root NARME as the name suggests contains two root of different orders or same orders wrapped around max and min functions. The job of the max function is to capture those values which are in the range $[1,\infty) \cup (-\infty, -1]$ and the min function captures values in the range $[0,1) \cup (-1,0)$.

This can be understood my the following example:-

Case 1: $\hat{y} = 1, y = 2$, then

$$max(\sqrt[n_t]{|4-2|}, 1) = \sqrt[n_t]{2} \text{ and}$$

$$min(\sqrt[m_t]{|4-2|}, 1) = 1$$

Here the max function captures the values

Case 2: $\hat{y} = 0.2, y = 0.6$, then

$$max(\sqrt[n_t]{|0.2-0.6|}, 1) = 1$$

$$min(\sqrt[m_t]{|0.2-0.6|}, 1) = \sqrt[m_t]{0.4}$$

Here the min function captures the values

Thus, together the max and min functions can cover values ranging from $(-\infty, \infty)$.

In the experiments it is found that for Single Root NARME even orders works well for most of the problems and for Double Root NARME even orders for first root and odd orders for the second root work well.

# 4 Experiments

### 4.1 Training a Neural Accumulator to add two numbers

A neural accumulator cell [1] is able to learn to add or subtract numbers and extrapolate that knowledge to values which were not present in the training set. The neural accumulator when trained using a single root NARME produced exceptional results, with accuracy upto 4 places of decimal in certain cases. Unlike other commonly used losses, the loss in case of a SR-NARME started from a much lower value which resulted in extremely fast training. While using Mean Squared Error, the model required 25,000 epochs to give fair results, but that model failed to accurately predict even one place of decimal, on the other hand the proposed loss function could predict at least upto 2 decimal places of precision in just 2,500 epochs, which is exactly one-tenth of that consumed taken by MSE.

A single root NARME of 10th order was used in this experiment, values greater than 10 are not found to be of much use. In this experiment only even values of $n_t$ worked find. As the order of the root increased the number of epochs required decreased as was discussed in earlier sections.

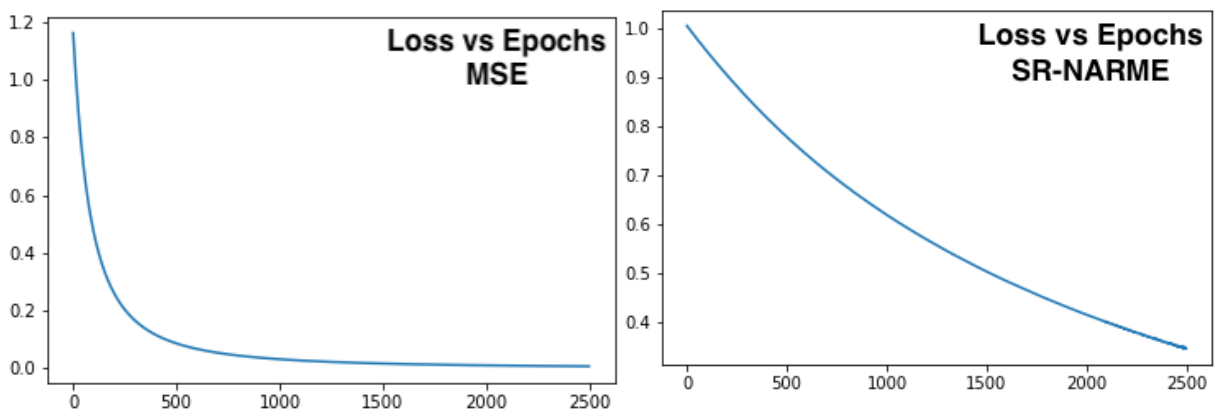

| S.No | Loss | Overall variance in test values |
|---|---|---|
| 1 | Huber | 1.15069646 |
| 2 | Log-cosh | 4.61058873 |
| 3 | MSE | 5.38001317 |
| 4 | MAE | 1.15070027 |
| 5 | SR-NARME | 0.00174397 |
| 6 | DR-NARME | 0.02953404 |

## 4.2 Training a Neural Arithmetic and Logic Unit to multiply two numbers

A neural arithmetic and logic unit cell [1] is an extension of a neural accumulator cell and can be used to learn to add, subtract, multiply, divide, etc using neural networks, like neural accumulator they too can extrapolate to values on which they were not trained. Using common regression losses it required 30,000 epochs to achieve acceptable accuracy, but when trained trained using a Double Root NARME it took only to 1470 epochs to get high accuracy (correctly predicted 8/10 product values), unlike using other regression losses which could give at most 6/10 correct values after training for 30,000 or even more number of epochs.

A DR-NARME of order 8 for first root and order 3 for second root was used in this experiment. It is found that for this particular experiment a double root NARME gives much better result than a single root NARME of order 8 and 10. In this case also the number of epochs and the n, m values were inversely proportional.

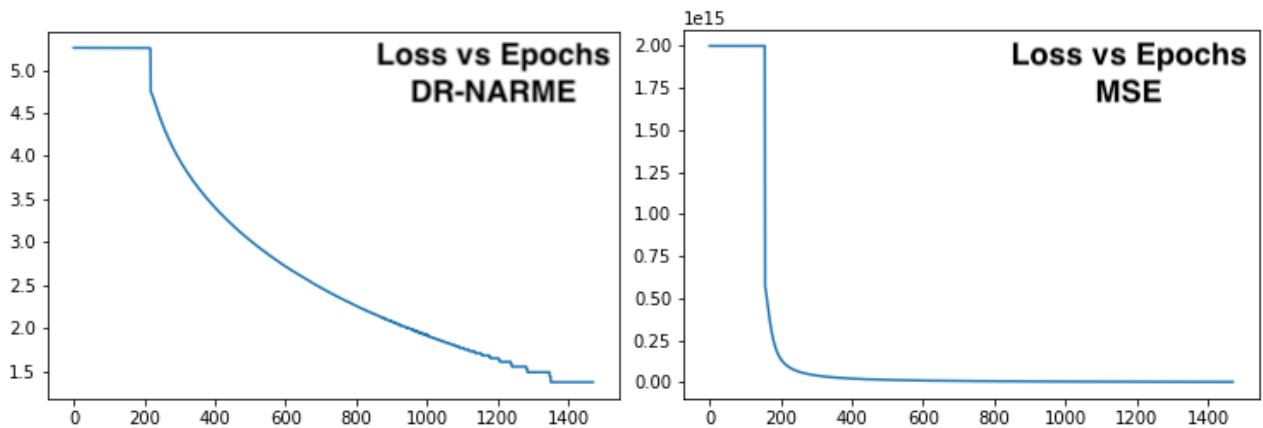

| S.No | Loss | Overall variance in test values |
|------|---------|-------|
| 1 | Huber | 29184 |
| 2 | Log-cosh | 45129 |
| 3 | MSE | 257566 |
| 4 | MAE | 45138 |
| 5 | SR-NARME | 4 |
| 6 | DR-NARME | 14 |

## 4.2 Predicting stock prices using an artificial neural network

Predicting stock prices requires the model to predict continuous values (regression task). As is evident from the graphs generated after training the model for 10 epochs, Double Root NARME performs the best for the same set of data inputs.

In this case a double root NARME of order 10 for first root and order 3 for second root worked very well. The overlapping of the predicted and original test data cannot be matched by other regression losses in such small number of training iterations.

A single root NARME also works well in this case. A SR-NARME of order 10 performs well compared to other losses but falls short when a double root is used instead. Like all other experiments in this experiment the number of epochs is inversely proportional to the order $n_t$, $m_t$ of the roots of the loss function.

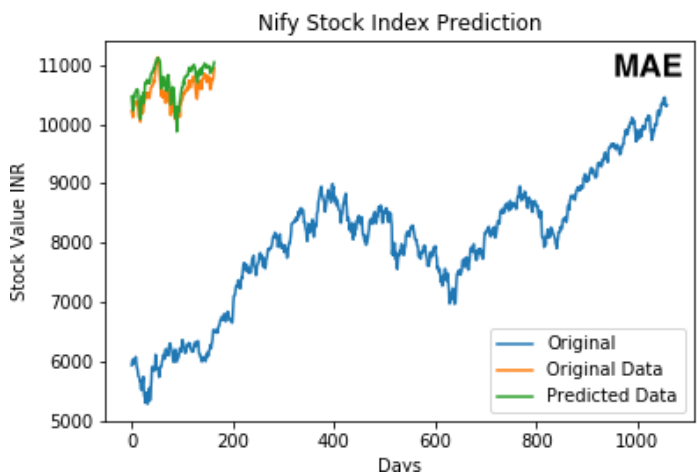
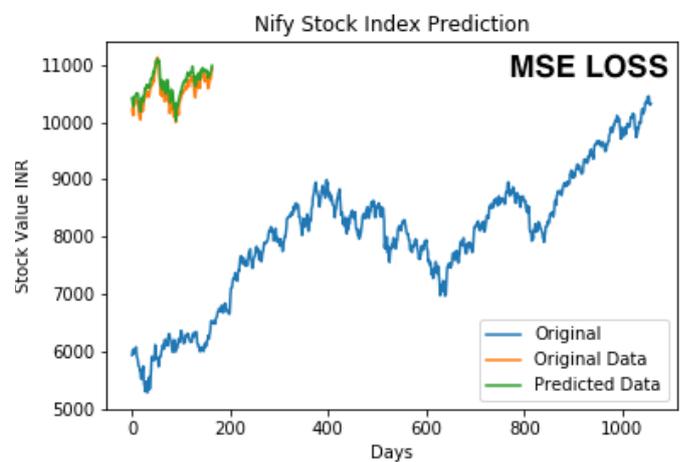

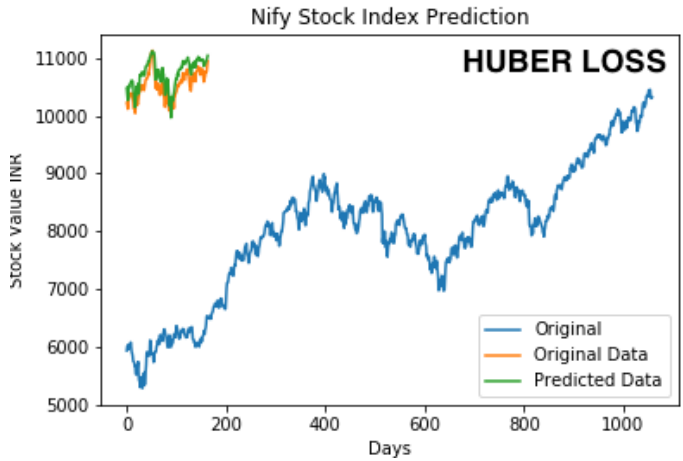
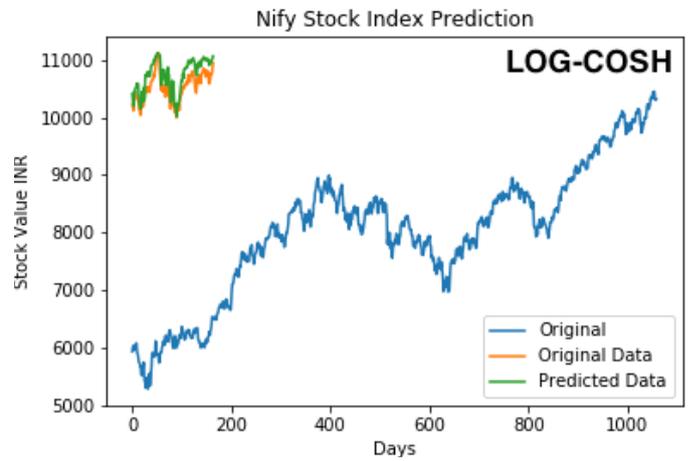
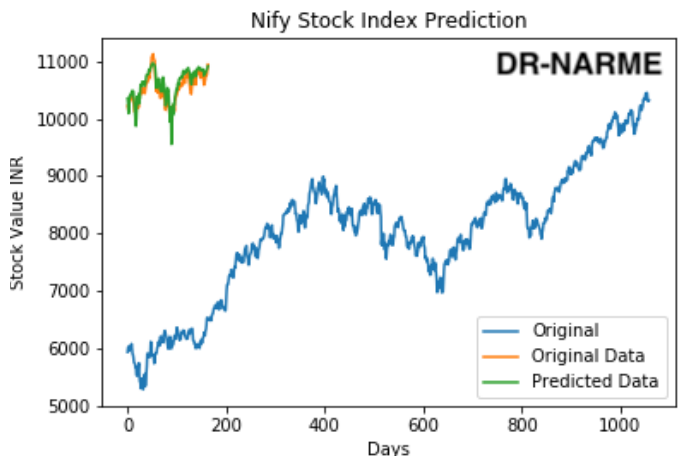

| S.No | Loss | Overall variance in test values |
|------|------|---------------------------------|
| 1 | Huber | 43922.97265913 |
| 2 | Log-cosh | 24132.882814 |
| 3 | MSE | 25216.63964994 |
| 4 | MAE | 26556.92871256 |
| 5 | SR-NARME | 22690.07617413 |
| 6 | DR-NARME | 20916.25097769 |

# 5 Conclusions

Current loss functions require a large amount of time when it comes to training models as compared to NARME which reduces the training time by a factor of almost ten. We have shown how NARME out performs some of the most commonly used loss functions for regression tasks in such small training time. It shows that using the Nth root of the absolute differences of the prediction and real values gives fast and accurate results. As is evident from the experiments that NARME greatly helps in the training process resulting in accuracies of upto 4 decimal places in some cases.

# 6 Acknowledgements

We would like to thank Andrew Ng for his deep learning specialization on Coursera which helped us start on the journey of deep learning.